\documentclass[]{ceurart} 
\usepackage{listings}
\usepackage{amsmath}
\usepackage{subcaption}

\begin{document}

\copyrightyear{2026} 
\copyrightclause{Copyright for this paper by its authors.
  Use permitted under Creative Commons License Attribution 4.0
  International (CC BY 4.0).}
\conference{Late-breaking work, Demos and Doctoral Consortium, colocated with the 4th World Conference on eXplainable Artificial Intelligence: July 01–03, 2026, Fortaleza, Brazil} 

\title{ConceptTracer: Interactive Analysis of Concept Saliency and Selectivity in Neural Representations}

\author[1]{Ricardo Knauer}[%
orcid=0009-0003-9258-5105,
email=ricardo.knauer@htw-berlin.de,
]
\cormark[1]

\author[1]{Andre Beinrucker}[%
email=andre.beinrucker@htw-berlin.de,
]

\author[1]{Erik Rodner}[%
email=erik.rodner@htw-berlin.de,
]

\address[1]{KI-Werkstatt, University of Applied Sciences Berlin, Berlin, Germany}
\cortext[1]{Corresponding author.}

\begin{abstract}
Neural networks deliver impressive predictive performance across a variety of tasks, but they are often opaque in their decision-making processes. Despite a growing interest in mechanistic interpretability, tools for systematically exploring the representations learned by neural networks in general, and tabular foundation models in particular, remain limited. In this work, we introduce ConceptTracer, an interactive application for analyzing neural representations through the lens of human-interpretable concepts. ConceptTracer integrates two information-theoretic measures that quantify concept saliency and selectivity, enabling researchers and practitioners to identify neurons that respond strongly to individual concepts. We demonstrate the utility of ConceptTracer on representations learned by TabPFN and show that our approach facilitates the discovery of interpretable neurons. Together, these capabilities provide a practical framework for investigating how neural networks like TabPFN encode concept-level information. ConceptTracer is available at \url{https://github.com/ml-lab-htw/concept-tracer}.
\end{abstract}


\maketitle

\section{Introduction}

Neural networks achieve remarkable predictive performance across a wide range of tasks, yet the mechanisms underlying their predictions are often difficult to interpret \citep{bommasani2022opportunitiesrisks,longo2024explainable}. This opacity can hinder their deployment in domains where transparency is critical \citep{adler2022deutsche,eu2024aiact}. Developing a deeper understanding of their learned representations can improve model interpretability and strengthen user trust, particularly in high-stakes environments such as clinical practice \citep{adler2022deutsche,eu2024aiact}. While mechanistic interpretability research has introduced advanced methods for revealing the inner workings of neural networks, \textit{e.g.}, by identifying interpretable units of computation \citep{ferrando2024primerinnerworkings,sharkey2025openproblems}, tools for systematically analyzing the representations of neural networks in general and tabular foundation models in particular remain scarce. As a result, researchers and practitioners often lack practical frameworks for examining how neural networks encode information and whether these representations align with higher-order, human-understandable features, \textit{i.e.}, concepts. Analyzing how such concepts are encoded within a model provides a principled way to investigate what information internal representations capture and to what extent they reflect human-interpretable structure.

Our contributions are as follows:

\begin{itemize}
    \item We introduce ConceptTracer, an interactive system for analyzing neural representations grounded in human-understandable concepts (Sect.~\ref{sec:concepttracer}). ConceptTracer incorporates two information-theoretic measures that quantify the neural saliency and selectivity for individual concepts \citep{hewitt2019designing,kandel2021principles,simonyan2014deepinsidecnns} (Sect.~\ref{sec:formal}).
    \item We demonstrate the practical value of ConceptTracer on representations learned by TabPFN \citep{grinsztajn2026tabpfn25,hollmann2025accurate}. Our results show that our approach supports the identification of interpretable neurons (Sect.~\ref{sec:experiments}).
\end{itemize}

\section{Related work}

A central question in mechanistic interpretability is whether individual neurons are directly interpretable because they respond strongly to single concepts, an idea linked to the ``grandmother cell'' hypothesis in neuroscience \citep{gross2002genealogy,quiroga2005invariant}. However, frameworks for analyzing neural representations at the level of individual neurons and concepts remain underexplored relative to established explainability tools \citep{dijk2026explainerdashboard,microsoft2026responsibleaidashboard} and recent methods that localize and disentangle distributed and superposed representations via auxiliary model training \citep{bertsimas2021sparse,gao2025scaling,gurnee2023finding,lieberum2024gemma,templeton2024scaling}. To the best of our knowledge, Neuronpedia \citep{neuronpedia} is currently the only application that supports an interactive exploration and visualization of neural representations, even though its functionality centers on features learned by sparse autoencoders and concepts identified through unsupervised discovery, which require substantial computational resources, rather than neurons in the underlying model and predefined concepts.

In the next section, we introduce two metrics to analyze neural representations at the level of individual model neurons and ground-truth concepts \citep{hewitt2019designing,kandel2021principles,simonyan2014deepinsidecnns} using information theory \citep{dayan2005theoretical,oikarinen2023clip,shannon1948mathematical}.

\section{Quantifying neural interpretability} \label{sec:formal}

In this section, we introduce our two information-theoretic measures for neural interpretability. Let $\mathbf{A} \in \mathbb{R}^{M \times N}$ be the neural activations for $M \in \mathbb{N_+}$ samples and $N \in \mathbb{N_+}$ neurons. The network structure is irrelevant for the following definitions and is thus omitted. Moreover, let $\mathbf{B} \in \{0, 1\}^{M \times C} $ represent the concept labels for the $M$ samples and $C \in \mathbb{N_+}$ higher-order, human-interpretable features, \textit{i.e.}, concepts. To quantify the association between the activations $\mathbf{a}_i \in \mathbb{R}^M$ of neuron $i$ and the concept labels $\mathbf{b}_j \in \{0, 1\}^M$ of concept $j$, we employ the normalized mutual information $\hat{I}(\mathbf{a}_i, \mathbf{b}_j) \in [0,1]$ as a nonlinear measure of statistical dependence \citep{dayan2005theoretical,oikarinen2023clip,shannon1948mathematical}. We define the saliency \citep{kandel2021principles, simonyan2014deepinsidecnns} for neuron $i$ with respect to concept $j$ as this mutual information:
\begin{align}
\text{saliency}(\mathbf{a}_i, \mathbf{b}_j) = \hat{I}(\mathbf{a}_i, \mathbf{b}_j), \quad 0 \leq \text{saliency}(\mathbf{a}_i, \mathbf{b}_j) \leq 1 \quad.
\label{eq:saliency}
\end{align}
The saliency thus captures the strength of association between a neuron and a concept. However, it does not indicate whether a neuron is specialized for a particular concept. To capture this specialization, we define the selectivity \citep{hewitt2019designing,kandel2021principles} as the fraction of a neuron’s total saliency that is attributable to a given concept:
\begin{align}
\text{selectivity}(\mathbf{a}_i, \mathbf{b}_j) = \frac{\text{saliency}(\mathbf{a}_i, \mathbf{b}_j)}{\sum_{c=1}^{C} \text{saliency}(\mathbf{a}_i, \mathbf{b}_c)}, \quad 0 \leq \text{selectivity}(\mathbf{a}_i, \mathbf{b}_j) \leq 1 \quad.
\label{eq:selectivity}
\end{align}
By construction, the selectivity depends on the size of the concept set: smaller concept sets tend to yield higher selectivity values due to the normalization. Moreover, the selectivity only accounts for the concepts $c\in \{1, ..., C\}$ under consideration. To determine whether our information-theoretic measures exceed what would be expected by chance, we employ nonparametric permutation testing \citep{franccois2006permutation} to estimate empirical null distributions for both metrics. Our null hypothesis $H_0$ is that the neural activations and concept labels are independent, \textit{i.e.}, no neuron is preferentially associated with any concept and no concept is preferentially associated with any neuron. Permutation tests are well suited in this setting because they only require that samples are exchangeable under permutations of the concept labels, while parametric tests typically rely on null distributions that are derived under much stronger assumptions \citep{knauer2026searchgrandmothercellstracing,ince2017statistical}. We generate permutations by applying a single shuffle of the sample indices across all concept labels, thereby preserving structural dependencies within the concept label space, and count the number of permutations in which the permuted test statistic equals or exceeds the observed value. This procedure yields empirical p-values for the saliency and the selectivity, $p_{saliency}(\mathbf{a}_i, \mathbf{b}_j)$ and $p_{selectivity}(\mathbf{a}_i, \mathbf{b}_j)$. To control the family-wise error rate, we apply a max-statistic correction \citep{westfall1993resampling} to both p-values and subsequently combine them into $p_{combined}(\mathbf{a}_i, \mathbf{b}_j)$ using a Bonferroni correction \citep{nikolitsa2025metacp}.

In the next section, we show how concept saliency and selectivity can be integrated into an interactive mechanistic interpretability dashboard to facilitate the discovery of interpretable neurons in neural representations.

\section{ConceptTracer} \label{sec:concepttracer}

\begin{figure}[t]
  \centering
  \includegraphics[width=0.9\textwidth]{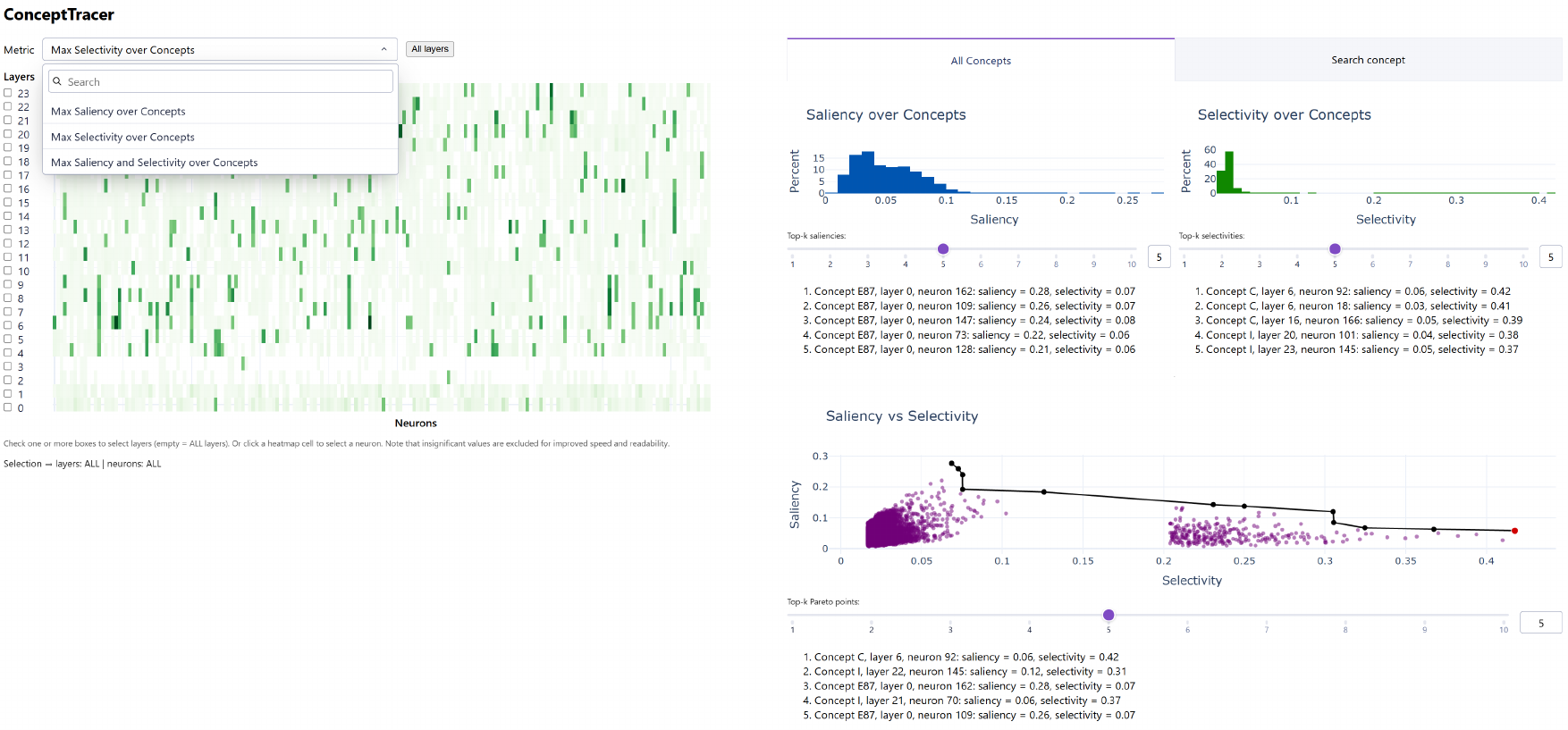}
  \caption{The ConceptTracer dashboard for the interactive analysis of neural representations.}
  \label{fig:dashboard}
\end{figure}

In this section, we introduce ConceptTracer\footnote{\url{https://github.com/ml-lab-htw/concept-tracer}}, an interactive application for analyzing neural representations through the lens of human-interpretable concepts that is built around four main components:
\begin{itemize}
    \item \textit{Dataset}: A set of training and test samples.
    \item\textit{Neural network}: A model for training and testing.
    \item\textit{Embedding extractor}: A function for extracting neural representations from the trained model.
    \item \textit{Concepts}: A set of higher-order, human-interpretable features.
\end{itemize}
Once the components are specified, ConceptTracer computes the concept saliency and selectivity, along with their individual and combined p-values, for any neuron-concept pair (Sect.~\ref{sec:formal}). These measures can then be systematically explored through the ConceptTracer dashboard (Fig.~\ref{fig:dashboard}). By default, only significant values at a significance level of 0.05 are displayed.

The dashboard consists of two main panels. The left panel enables users to explore the maximum saliency, selectivity, or combined score across concepts at different levels of the network: network-wide, per layer, or per neuron. The right panel visualizes the distribution of the metrics within the chosen subnetwork and highlights the top-k neuron-concept pairs. To facilitate the identification of pairs that are both salient and selective, the Pareto front is displayed. In addition, the knee point on the Pareto front is marked to indicate the most salient and selective pair. The knee point is determined by maximizing the sum of the min-max scaled saliency and selectivity scores. Beyond concept-wide exploration, users can also search for specific concepts and perform analyses at the level of individual concepts through the \texttt{Search concept} tab. ConceptTracer therefore enables the examination of neuron-concept associations from four complementary perspectives:
\begin{itemize}
    \item \textit{Network view}: How salient and selective are the neurons across the entire network for the set of concepts?
    \item \textit{Layer view}: How salient and selective are the neurons within a selected layer or set of layers for the set of concepts?
    \item \textit{Neuron view}: How salient and selective is a selected neuron for the set of concepts?
    \item \textit{Concept view}: How salient and selective are the neurons in the network, selected layer, or set of layers for a particular concept?
\end{itemize}
This allows for a comprehensive investigation of how neural networks encode concept-level information. In the next section, we empirically test whether we can use ConceptTracer to find interpretable neurons in neural representations.

\section{Experiments} \label{sec:experiments}

In this section, we use our definitions of concept saliency and selectivity (Sect.~\ref{sec:formal}) within ConceptTracer (Sect.~\ref{sec:concepttracer}) to evaluate whether our framework can uncover interpretable neurons in neural representations. To this end, we apply ConceptTracer on representations learned by TabPFN \citep{grinsztajn2026tabpfn25,hollmann2025accurate} and demonstrate that our approach supports the identification of interpretable neurons.

\subsection{Experimental setup} \label{sec:setup}

\subsubsection{Dataset and tasks}

Understanding model behavior is particularly important for mitigating risks in high-stakes domains such as clinical practice \citep{adler2022deutsche,eu2024aiact}. Therefore, we conducted our experiments using the MIMIC-IV-ED dataset to solve four emergency department (ED) triage tasks: inhospital mortality prediction, intensive care unit (ICU) transfer within 12h prediction, critical outcome prediction  (defined as either inhospital mortality or ICU transfer within 12h), and hospitalization prediction \citep{xie2022benchmarking}. Each task was treated as a binary classification problem. The data was collected at the Beth Israel Deaconess Medical Center between 2011 and 2019 and encompassed 64 features, spanning patient history, demographics, vital signs, chief complaints, and comorbidities. The training set contained 353,150 ED episodes (samples) from 182,588 unique patients, the test set 88,287 ED episodes from 65,169 unique patients \citep{xie2022benchmarking}. Concepts were defined using diagnostic codes from the International Classification of Diseases (ICD) organized at three hierarchical levels:
\begin{itemize}
    \item \textit{High-level}: \textit{e.g.}, R = signs and symptoms.
    \item \textit{Mid-level}: \textit{e.g.}, R57 = shock.
    \item \textit{Low-level}: \textit{e.g.}, R570 = cardiogenic shock.
\end{itemize}

\subsubsection{Data and concept preprocessing}

We excluded training set patients who also appeared in the test set to prevent data leakage and removed rows without available ICD codes. For all tasks except hospitalization prediction, the minority class constituted $\leq 8\%$ of the training samples. To address the class imbalance, we undersampled the majority classes in the training sets, resulting in 352, 1,318, 1,498, and 7,512 training samples for mortality, ICU transfer, critical outcome, and hospitalization prediction, respectively. For the low- and mid-level concepts, the concept prevalence followed a long-tailed distribution, with a large fraction of concepts supported by very few or even single samples. To mitigate the concept imbalance, we filtered out concepts with a prevalence $< 100$, yielding 354, 282, and 24 low-, mid-, and high-level concepts, respectively.

\subsubsection{Model and evaluation}

We employed TabPFN 6.3.2 \citep{grinsztajn2026tabpfn25,hollmann2025accurate} as our neural network due to its strong reported performance in tabular prediction tasks \citep{erickson2025tabarena}. TabPFN was used without tuning or ensembling and its discriminative performance was evaluated using the test set area under the receiver operating characteristic curve (AUC). Next, we extracted the 192-dimensional target-column embeddings \citep{hollmann2025accurate} from each of its 24 encoder layers \citep{grinsztajn2026tabpfn25} and passed them, together with the concept labels, to ConceptTracer to systematically analyze the neuron-concept pairs. As baselines, we employed sparse probes based on SHAP values \citep{covert2021explaining,lundberg2017unified} and optimal probing \citep{bertsimas2021sparse,gurnee2023finding}. Please refer to \citet{knauer2026searchgrandmothercellstracing} for implementation details on the baselines.

\subsection{Experimental results}

\begin{figure}[t]
  \centering
  \begin{subfigure}{0.47\textwidth}
    \includegraphics[width=\textwidth]{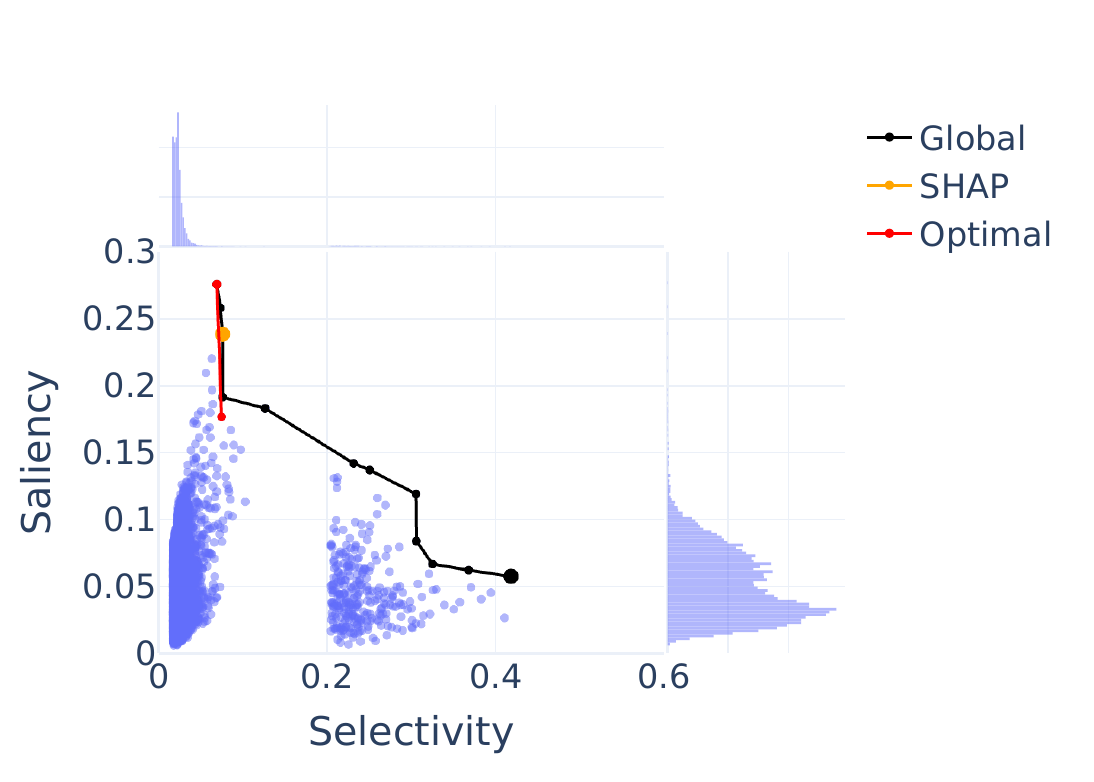}
    \caption{Inhospital mortality.}
  \end{subfigure}
  \begin{subfigure}{0.47\textwidth}
    \includegraphics[width=\textwidth]{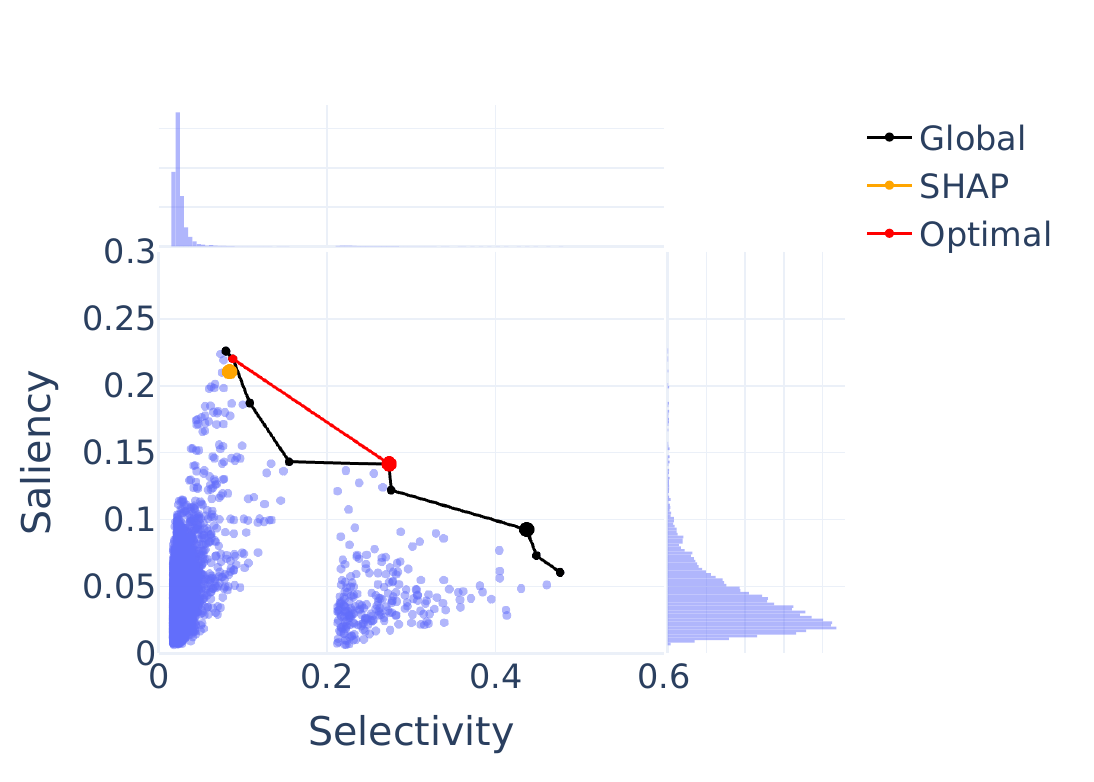}
    \caption{ICU transfer within 12h.}
  \end{subfigure}
  \begin{subfigure}{0.47\textwidth}
    \includegraphics[width=\textwidth]{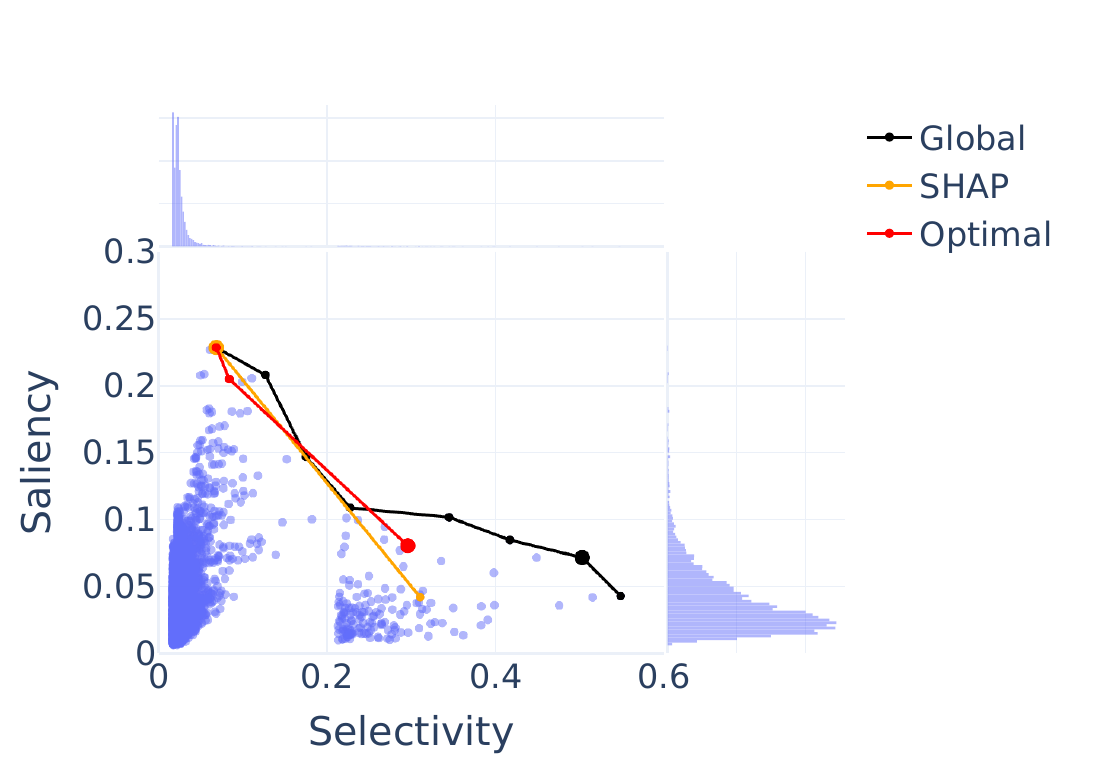}
    \caption{Critical outcome.}
  \end{subfigure}
  \begin{subfigure}{0.47\textwidth}
    \includegraphics[width=\textwidth]{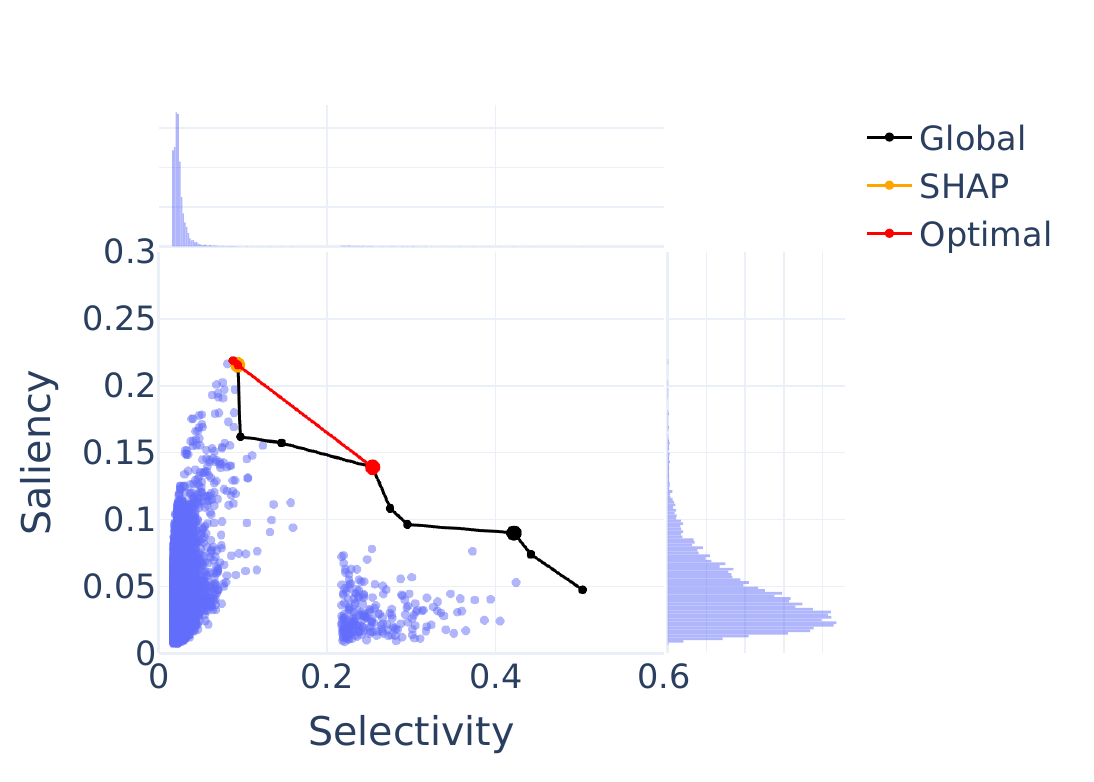}
    \caption{Hospitalization.}
  \end{subfigure}
  \caption{Significant neuron-concept pairs for our tasks. Black denotes the global Pareto front, with larger markers indicating knee points. The Pareto fronts for the sparse probing baselines via SHAP values and optimal probing are shown in orange and red, respectively.}
  \label{fig:results}
\end{figure}

TabPFN achieved test set AUC values ranging from 0.79 to 0.85. Less than $1\%$ of the neurons were significantly salient and selective for specific concepts. Due to the normalization in the selectivity calculation (Eq.~\ref{eq:selectivity}), the significant neuron-concept pairs formed two distinct clusters of low- and mid-level concepts with smaller selectivity values versus high-level concepts with larger selectivity values (Fig.~\ref{fig:results}). Sparse probes via SHAP values and optimal probing only partially recovered the global Pareto fronts and missed the most selective pairs (Fig.~\ref{fig:results}). This suggests that identifying the most salient and selective neuron-concept pairs requires an exhaustive search rather than relying solely on sparse probing approaches.

The global Pareto fronts included neuron-concept pairs from initial, middle, and final layers of TabPFN, indicating that concepts may not be preferentially represented in particular layers. The neurons on the global Pareto fronts were associated with the following concepts:
\begin{itemize}
    \item \textit{Inhospital mortality}: C (malignant neoplasms), E66 (obesity), E87 (other disorders of fluid, electrolyte, and acid-base balance), I (diseases of the circulatory system).
    \item \textit{ICU transfer within 12h}: F (mental and behavioral disorders), F32 (depressive episode).
    \item \textit{Critical outcome}: F, F32.
    \item \textit{Hospitalization}: F, F32, I50 (heart failure).
\end{itemize}
These results suggest two complementary neural response patterns that are consistent with concept differentiation within composite input features and concept integration across related input features. For example, neuron 162 in layer 0 on the Pareto front shows salient and selective responses to ICD code E87 (other disorders of fluid, electrolyte, and acid-base balance). In contrast, its saliency and selectivity for other subconcepts within the fluid and electrolyte disorders feature are substantially reduced, \textit{e.g.}, by 0.12 and 0.03 for ICD code E86 (volume depletion). This pattern is consistent with differentiation within a composite input feature. Conversely, neuron 145 in layer 22 on the Pareto front shows salient and selective responses to ICD code I (diseases of the circulatory system). When conditioned on the presence of ICD code I, cardiac arrythmias and hypertension rank among the top-4 features by mutual information with the neuron's activations. This pattern is consistent with integration across related input features.

Overall, our results demonstrate that ConceptTracer facilitates the discovery of interpretable neurons and provides a framework for systematically exploring how neural networks encode concept-level information.

\section{Conclusion}

Interpreting the decision-making processes of neural networks remains a big challenge. To address this challenge, we introduce ConceptTracer, an interactive application for analyzing neural representations by relating them to human-interpretable concepts. We show that ConceptTracer supports the identification of interpretable neurons in the tabular foundation model TabPFN and facilitates deeper exploration of how the network encodes concept-level information.

Several promising directions emerge from our work. For a more fine-grained analysis, future work could extend ConceptTracer by decomposing the association between neural representations and concepts into unique, redundant, and synergistic components via partial information decomposition \citep{williams2010pid}. For bias detection and mitigation, our framework could be applied to find bias-related neurons and reduce their influence via activation steering \citep{dev2020measuring}. Finally, ConceptTracer could help to better understand and deliberately shape the inductive biases of tabular foundation models toward meaningful real-world concepts \citep{grinsztajn2026tabpfn25}. We believe that our framework provides a valuable resource for practitioners and researchers working in mechanistic interpretability.

\begin{acknowledgments}
This research was funded by the European Union’s Erasmus+ program (project: EUonAIR, project number: 101177370) and the Deutsche Forschungsgemeinschaft (DFG, German Research Foundation, project: Berlin Initiative for Applied Foundation Model Research, project number: 528483508).
\end{acknowledgments}

\section*{Declaration on Generative AI}
During the preparation of this work, the authors used GPT-5 for dashboard prototyping. After using these tools/services, the authors reviewed and edited the content as needed and take full responsibility for the content. 

\bibliography{sample-ceur}

\end{document}